\documentclass[10pt,twocolumn,letterpaper]{article}

\usepackage[pagenumbers]{cvpr}
\usepackage{tabularx}
\usepackage{multirow}
\usepackage{array}

\usepackage{microtype}
\usepackage{tcolorbox}
\tcbset{
  agentprompt/.style={
    boxrule=0.45pt,
    arc=2pt,
    left=5pt,
    right=5pt,
    top=4pt,
    bottom=4pt,
    before skip=4pt,
    after skip=7pt,
    fonttitle=\bfseries\small,
    fontupper=\small\ttfamily,
    before upper={\raggedright}
  }
}

\usepackage{enumitem}
\setlist{topsep=2pt,partopsep=0pt,itemsep=2pt,parsep=0pt,leftmargin=1.4em}

\definecolor{cvprblue}{rgb}{0.21,0.49,0.74}
\PassOptionsToPackage{hyphens}{url}
\usepackage[breaklinks,colorlinks,allcolors=cvprblue]{hyperref}

\def\paperID{*****}
\def\confName{CVPR}
\def\confYear{2026}

\newcommand{\vlmalong}{Perceptual Abstraction Engine\xspace}
\newcommand{\vlmashort}{PAE\xspace}
\newcommand{\vlmblong}{Cognitive Retrieval Engine\xspace}
\newcommand{\vlmbshort}{CoRE\xspace}
\newcommand{\vlmclong}{Multimodal Executive Controller\xspace}
\newcommand{\vlmcshort}{MEC\xspace}
\newcommand{\memolong}{Episodic Visual Memory\xspace}
\newcommand{\memoshort}{EVM\xspace}
\newcommand{\agentlong}{Cognitive-structured Multimodal Agent\xspace}
\newcommand{\cmaharness}{CMA-Harness\xspace}
\newcommand{\cmaharnesslong}{\agentlong Harness\xspace}
\newcommand{\dataengine}{Unified Scenario Engine\xspace}

\newcommand{\benchlong}{Multi-turn Context Agent Benchmark\xspace}
\newcommand{\benchshort}{M2CA-Bench\xspace}

\begin{document}

\title{\agentlong for Multimodal Understanding, Generation, and Editing}

\author{Feng Wang$^{1}$\thanks{Work done during an internship at WeChat Vision, Tencent Inc.}, Canmiao Fu$^{2}$, Zhipeng Huang$^{2}$, Chen Li$^{2}$, Jing LYU$^{2}$, Ge Li$^{1}$\\
$^{1}$Peking University \quad $^{2}$WeChat Vision, Tencent Inc.}

\maketitle

\begin{abstract}

Recent unified multimodal models demonstrate that a single architecture can jointly perform vision/language understanding and image generation/editing. However, these monolithic designs rely on repeatedly feeding all historical visual and textual inputs into a shared context window, limiting scalability in long-horizon multimodal dialogue due to visual token explosion and unreliable cross-turn visual referencing. In this work, we propose a \agentlong that externalizes visual information into an Episodic Visual Memory and selectively reactivates relevant visual episodes during reasoning. The agent consists of a \vlmalong for structured visual abstraction, a \vlmblong for cross-turn memory retrieval, and a \vlmclong for autonomous task inference and action planning. To address the lack of turn-level retrieval supervision in existing multimodal dialogue datasets, we further develop a \dataengine that programmatically generates structured multi-turn conversations with fine-grained retrieval annotations, enabling reinforcement learning to optimize perceptual abstraction and retrieval policies. We additionally construct a long-horizon visual-dialogue benchmark and stratify it by difficulty to evaluate episodic visual recall. Extensive experiments show that our 8B agent achieves 91.4\% retrieval accuracy over 20-turn sessions, surpassing 32B baselines by +8.2\%, while nearly halving per-turn inference time (23.1s $\rightarrow$ 12.7s). We further present the \cmaharnesslong (\cmaharness), a tool-augmented deployment of the same cognitive structure that integrates persistent multimodal memory, web access, image generation/editing/composition tools, and OpenAI-compatible serving. These results suggest that structured memory and modular decision-making provide a more scalable and efficient paradigm for long-horizon multimodal agents than monolithic parameter scaling. Our \href{https://github.com/caseclose/cma-harness}{code}, dataset, and project page (\href{https://caseclose.github.io/cma-harness/}{caseclose.github.io/cma-harness}) will all be released.

\end{abstract}

\section{Introduction}
\label{sec:intro}

\begin{figure*}[t]
  \centering
  \IfFileExists{image/dialogue_0448.pdf}{\includegraphics[width=\textwidth]{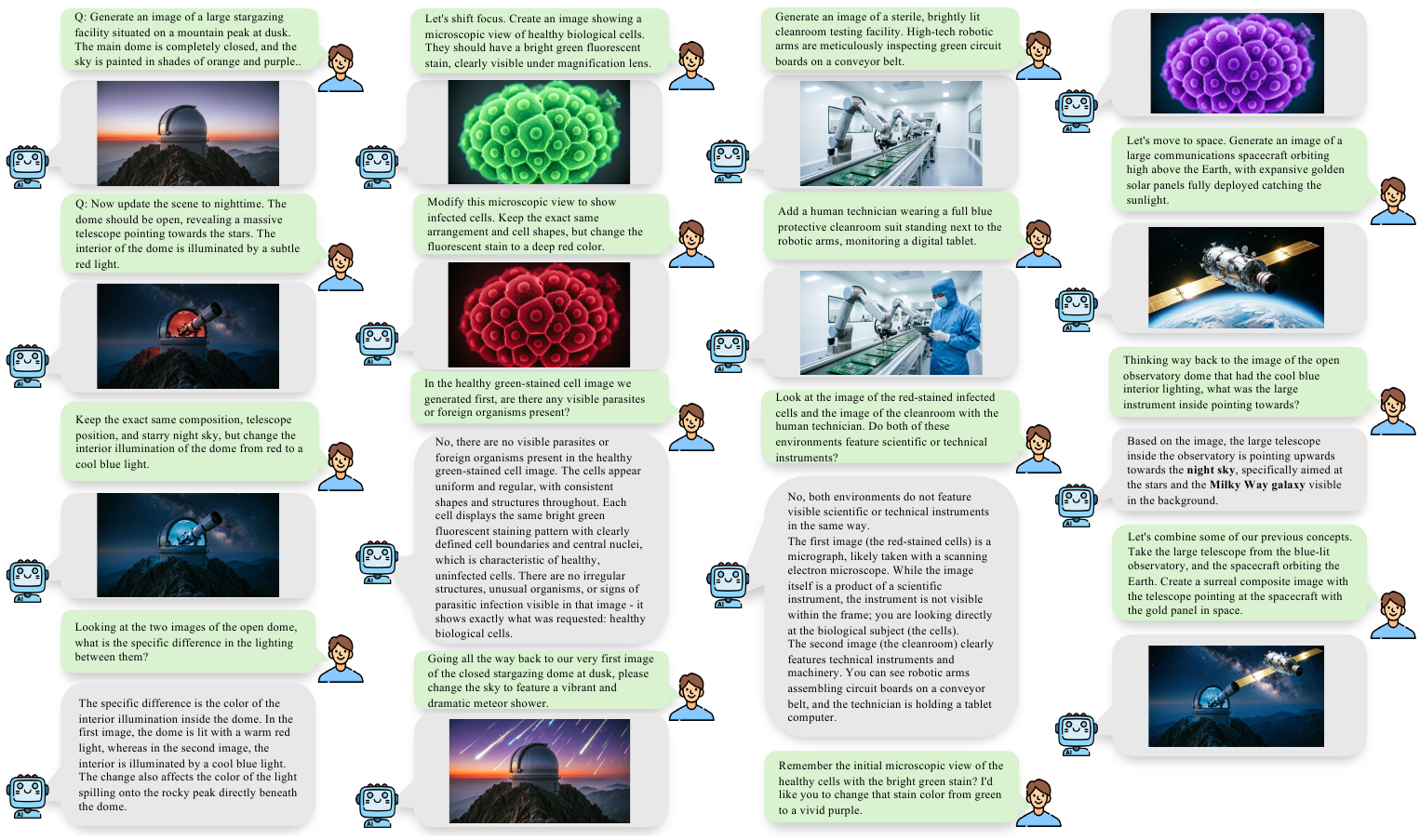}}{\fbox{\parbox{0.92\textwidth}{Overview dialogue placeholder: image/dialogue\_0448.pdf is not included in the source tree.}}}
  \caption{A multi-turn multimodal dialogue produced by our \agentlong, spanning 20 turns across four topics (stargazing dome, biological cells, cleanroom facility, spacecraft). The agent autonomously handles interleaved understanding, generation, and editing tasks while accurately retrieving and referencing visual episodes from earlier turns.}
  \label{fig:overview_dialogue}
\end{figure*}

Unified multimodal models have recently shown that a single architecture can jointly perform vision-language understanding and image generation and editing. Recent approaches~\cite{ChameleonTeam2024,wang2024emu3,Xie2025Showo2,Ma2025JanusFlow,Deng2025BAGEL,huang2025wegen} consolidate perception and generation within a shared parameter space and achieve strong performance across diverse tasks. These systems typically treat multimodal interaction as autoregressive prediction over a unified token sequence.

While effective for short-context interaction, this unified paradigm exhibits structural limitations in long-horizon multimodal dialogue. In practical image-text conversations, users frequently reference images introduced many turns earlier, request iterative modifications, or switch between understanding and generation without explicit task indicators. Figure~\ref{fig:overview_dialogue} shows one such session, spanning 20 turns and four distinct topics. Under such settings, repeatedly injecting all historical visual tokens into the context window leads to two major issues. First, visual tokens are substantially more expensive than text tokens; as dialogue length increases, token usage grows rapidly and crowds out the context budget available for reasoning. Second, reliance on implicit attention over extended contexts weakens cross-turn visual referencing, resulting in retrieval errors and semantic drift. These challenges indicate that parameter scaling alone is insufficient for sustained multimodal interaction. Indeed, a strong unified model (BAGEL~\cite{Deng2025BAGEL}) retrieves the correct visual episode in fewer than 3\% of the hardest turns of our benchmark (Sec.~\ref{subsec:Accuracy_for_context_usage}), despite excelling at single-turn tasks.

Recent agent-based approaches attempt to address parts of this problem. Memory-augmented video agents~\cite{fan2024videoagent,Yeo2025WorldMM} maintain episodic buffers for long video reasoning, and multi-agent generation systems~\cite{wang2024genartist,Venkatesh2025CREA} coordinate specialized modules for image creation and editing. However, these methods either operate on a single input stream (video) or a single task family (image creation), and none simultaneously addresses cross-turn visual retrieval, mixed-task orchestration, and long-horizon context management within a unified dialogue framework. Text-memory agents~\cite{ouyang2025reasoningbank,ye2025agentfold} externalize long-term state into semantic memory banks, but their purely textual abstractions discard the fine-grained visual detail needed to disambiguate near-duplicate images (Sec.~\ref{subsec:memory_representation_ablation}).

\begin{figure*}[t]
  \centering
  \IfFileExists{image/pipeline_new.png}{\includegraphics[width=0.98\textwidth]{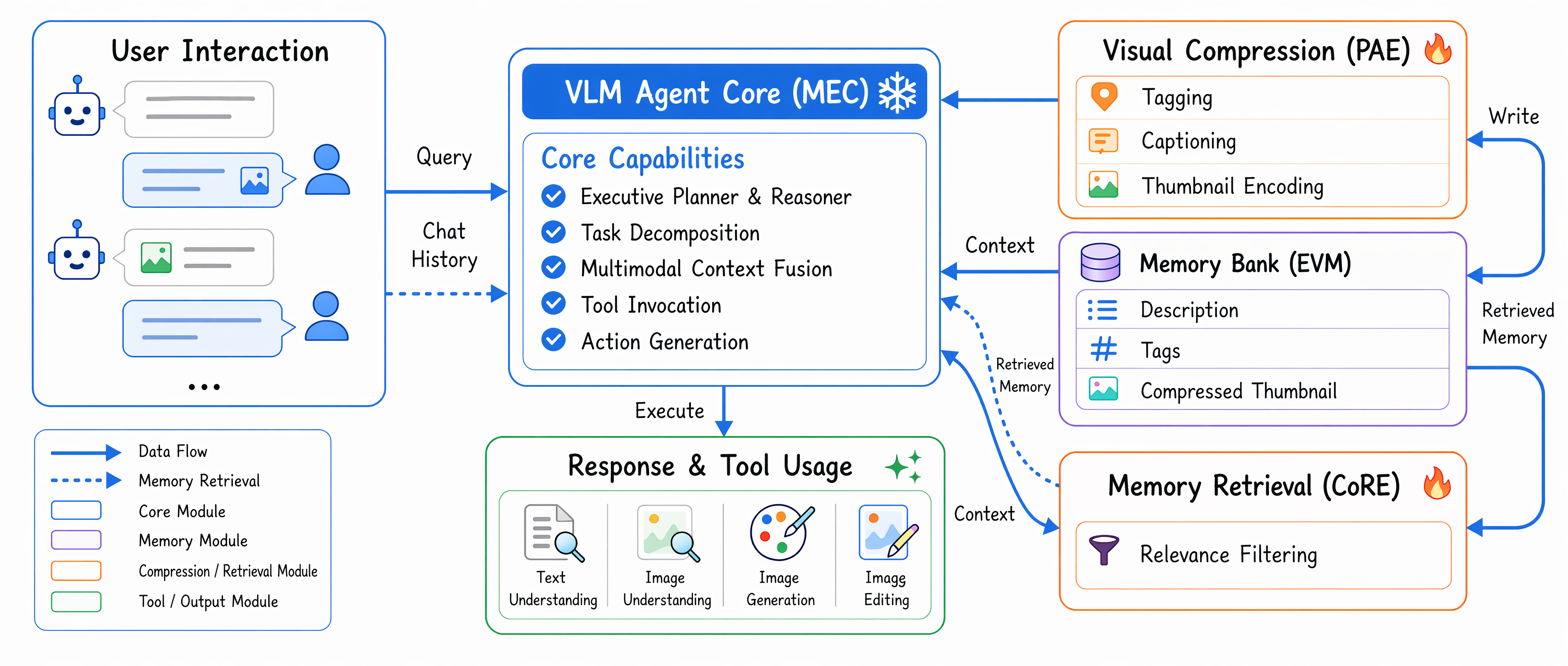}}{\fbox{\parbox{0.92\textwidth}{Pipeline placeholder: image/pipeline\_new.png is not included in the source tree.}}}
  \caption{End-to-end pipeline of the proposed \agentlong. Each incoming image passes through the \vlmalong (\vlmashort), which compresses it into a structured entry (tags, description, thumbnail) written to the \memolong (\memoshort). Given the current query and dialogue state, the \vlmblong (\vlmbshort) selects only the visual episodes relevant to this turn. The \vlmclong (\vlmcshort) integrates the retrieved memory with the dialogue context and dispatches the request to understanding, generation, editing, composition, or pure chat, keeping per-turn visual-token consumption bounded regardless of dialogue length.}
  \label{fig:pipeline}
\end{figure*}

Building a long-horizon multimodal agent therefore requires explicit memory management and autonomous decision-making: rather than treating images as persistent tokens, visual information should be externalized into structured memory units and selectively reactivated on demand. As Fig.~\ref{fig:pipeline} shows, we propose a \textbf{\agentlong} that decouples perception, episodic memory, retrieval reasoning, and executive control. Incoming images are first processed by a \textbf{\vlmalong (\vlmashort)}, which produces structured semantic abstractions including descriptive captions, attribute tags, and compact thumbnails. These representations are stored in an external \textbf{\memolong (\memoshort)}, allowing visual content to persist without occupying the dialogue context window. Given the evolving dialogue state, a \textbf{\vlmblong (\vlmbshort)} performs cross-modal reasoning to retrieve relevant visual episodes. A \textbf{\vlmclong (\vlmcshort)} integrates dialogue context and retrieved memory, autonomously infers task intent (understanding, generation, editing, composition, or pure chat), and plans the final action.

By retrieving only relevant visual episodes, the agent avoids repeatedly feeding all historical images into the model, significantly reducing visual token overhead while maintaining long-range consistency. Decoupling the executive controller (\vlmcshort) from the retrieval engine (\vlmbshort) further lets each module be trained, replaced, or scaled independently, so that model capacity can be allocated where it matters most for the performance--latency trade-off.

However, developing such memory-aware agents introduces a new challenge: existing multimodal dialogue datasets rarely provide explicit supervision for cross-turn visual retrieval. Most datasets~\cite{Wang2024MMNeedle,Wu2024VisualHaystacks} focus on single-turn grounding or short-context reasoning, and therefore lack fine-grained annotations indicating which historical visual episode should be recalled at each dialogue turn. To address this limitation, we introduce a \textbf{\dataengine} that programmatically generates structured multi-turn multimodal conversations. The engine simulates diverse visual scenes and dialogue trajectories, automatically producing fine-grained turn-level retrieval annotations that specify which visual episode is relevant to each query. Using these data, we apply a staged SFT+RL pipeline: the \vlmblong is first trained to retrieve, and the \vlmalong is then optimized against a \emph{frozen} retriever, so that memory representations are rewarded directly by downstream retrieval success rather than by captioning similarity.

To systematically evaluate long-horizon visual memory and agent behavior, we additionally construct the \textbf{\benchlong (\benchshort)}, consisting of 100 sessions of 20 turns each (2,000 annotated turns) with tasks randomly interleaved within each session. The benchmark stratifies turns into four difficulty levels according to temporal span, topic-shift frequency, multi-image interaction, and referential ambiguity, and injects hard negatives---high-similarity confounder images and no-retrieval-needed queries---to prevent shortcut learning. Unlike existing multimodal benchmarks that focus on single-turn reasoning, it specifically targets episodic memory retrieval in sustained multimodal interaction.

Extensive experiments show that our 8B agent reaches \textbf{91.4\%} retrieval accuracy over 20-turn English sessions (and 89.6\% in Chinese), surpassing the strongest 32B agent baseline by \textbf{+8.2\%} while nearly halving per-turn inference time (23.1s $\rightarrow$ 12.7s). The margin widens with dialogue horizon, growing from +9.6\% on the Full subset to +18.4\% on the Hard subset, while our approach maintains substantially lower visual token usage. Beyond the benchmark-optimized agent, we further instantiate the same cognitive structure as \cmaharness, which combines persistent multimodal memory with web search, image generation, editing, composition, and OpenAI-compatible serving. These results indicate that structured memory and modular decision-making provide a more scalable foundation for multimodal agents than monolithic parameter scaling.

To summarize, our main contributions are as follows: (1) We propose a \agentlong with explicit episodic visual memory for long-horizon dialogue that enables autonomous task inference. (2) We develop a \dataengine that programmatically generates structured multimodal conversations with turn-level retrieval supervision. (3) We construct a benchmark specifically designed to evaluate cross-turn visual retrieval in multimodal interaction. (4) We introduce reinforcement learning-based optimization techniques for perceptual abstraction and retrieval, enabling an 8B agent to surpass unified 32B baselines in long-range multimodal dialogue. (5) We present \cmaharness, a tool-augmented deployment of the same architecture that extends the memory-centric agent to open-ended multimodal workflows.

\section{Related Work}
\label{sec:related_work}

\subsection{Agents}
\label{sec:related_agents}

\noindent\textbf{LLM \& VLM Agents.}
Large Language Models~\cite{achiam2023gpt,touvron2023llama,bai2023qwen,qwen3} and Vision-Language Models~\cite{liu2024llavanext,bai2025qwen3vl,wang2025internvl3,peng2023kosmos,li2023blip,zhu2023minigpt} have driven autonomous agents capable of deliberative reasoning and tool execution~\cite{schick2024toolformer,Li2025SPORT}. To address context-window limits, recent agents incorporate explicit memory mechanisms~\cite{jin2025videomem,ouyang2025reasoningbank,ye2025agentfold} that externalize information into episodic and semantic memory banks~\cite{Jiang2026SYNAPSE,Yeo2025WorldMM}, mitigating the cost of attending to long histories. In multimodal domains, VLM agents~\cite{long2025seeing,kumar2024mmctagent,wang2025yanyun,liu2025agent0} act as controllers that orchestrate visual tools for long-form tasks such as video understanding~\cite{fan2024videoagent,wang2024videoagent}.

\noindent\textbf{Single \& Multi-agents.}
While single-agent systems~\cite{wang2024genartist,gao2024multi} excel at routing sub-tasks to expert tools with self-correction, multi-agent frameworks~\cite{li2023camel,Venkatesh2025CREA,hong2023metagpt,qian2024chatdev} distribute complex goals---such as creative generation---across specialized roles using programmable, collaborative protocols. However, existing visual agents typically operate within short contexts or restrict their scope to isolated domains. Crucially, our work introduces the first long-horizon, multi-agent architecture that seamlessly unifies pure text dialogue, multimodal understanding, image generation, and image editing within a single conversational framework. By leveraging an external episodic visual memory and a decoupled retrieval engine, our system maintains cross-turn intent and visual consistency without the exponential token costs of standard context scaling.

\subsection{Multimodal Models}
\label{sec:related_multimodal}

Foundational VLMs such as the Qwen-VL~\cite{Bai2023QwenVLAV} family have progressed from contrastive representation learning to instruction-following assistants capable of complex reasoning. Unified multimodal models~\cite{Deng2025BAGEL,tang2025unilip} consolidate perception and generation into a shared parameter space, either via token-based autoregression over interleaved text and visual tokens~\cite{ChameleonTeam2024,wang2024emu3,Deng2025BAGEL} or via hybrid architectures that blend autoregression with continuous diffusion or rectified flow~\cite{Xie2025Showo2,Ma2025JanusFlow}. In parallel, diffusion-based frameworks like InstructPix2Pix~\cite{brooks2023instructpix2pix} and TalkPhoto~\cite{Hu2026TalkPhoto} enable instruction-driven editing, with recent methods addressing error accumulation in multi-turn iterative editing~\cite{Zhou2025MultiTurn}.

Despite their elegance, scaling the context window is insufficient for long-horizon interactions: benchmarks such as MMNeedle~\cite{Wang2024MMNeedle} and Visual Haystacks~\cite{Wu2024VisualHaystacks} reveal that implicit attention over massive visual histories leads to severe retrieval degradation and semantic drift, and repeatedly encoding visual tokens in unified architectures is computationally exorbitant. By externalizing visual history into structured episodes, our framework bypasses the inherent token limits of unified models, effectively maintaining reliable, long-range visual grounding across interleaved understanding, generation, and editing tasks.

\section{Methods}
\label{sec:methods}

\subsection{Problem Formulation}
We formalize long-horizon multimodal dialogue as an interaction trajectory:
\[
  \mathcal{D} = \left\{ (u_i, x_i, y_i) \right\}_{i=1}^{T},
\]
where $u_i$ denotes the user text input at turn $i$, $x_i$ the optional visual input, and $y_i$ the system response. The dialogue history before turn $i$ is $\mathcal{H}_{<i} = \{(u_j, x_j, y_j)\}_{j=1}^{i-1}$, and the historical image set is $\mathcal{I}_{<i} = \{x_j\}_{j=1}^{i-1}$.

A naive approach feeds the full history into a multimodal model, $y_i = f(u_i, x_i, \mathcal{H}_{<i})$, which suffers from three limitations as $i$ grows: (i) prohibitive visual-token growth, (ii) contextual interference from irrelevant images, and (iii) unstable reasoning over long histories.

While text history carries essential semantic continuity, only a small \emph{subset} of past images is relevant at each turn---a fundamental asymmetry overlooked by existing unified models. We reformulate the problem as \textbf{visual-context selection}, learning a policy $\pi$ that selects a minimal relevant image subset and predicts from it:
\[
\mathcal{I}_i^* = \pi(u_i, \mathcal{H}_{<i}) \subseteq \mathcal{I}_{<i}, \qquad y_i = f\big(u_i, \operatorname{Text}(\mathcal{H}_{<i}), \mathcal{I}_i^*\big).
\]
Our objective is to learn $\pi$ that minimizes $|\mathcal{I}_i^*|$ while preserving downstream task performance. The \agentlong framework (Sec.~\ref{subsec:architecture}) implements $\pi$ via coordinated specialized agents.

\subsection{\agentlong}
\label{subsec:architecture}

Relying on a single monolithic model to compress memory, select context, and execute tasks results in conflicting objectives and unbounded visual-token growth. This limitation is confirmed empirically by the near-zero retrieval accuracy of unified baselines on the hardest turns (Sec.~\ref{subsec:Accuracy_for_context_usage}).

We therefore decompose the pipeline into the \textbf{\agentlong} framework (Fig.~\ref{fig:pipeline}), which assigns each sub-problem to a dedicated specialist, making every component independently trainable and upgradable:

\begin{itemize}
    \item \textbf{\vlmalong (\vlmashort)}: abstracts each incoming image into a structured memory entry, including semantic tags, textual description, and a compressed thumbnail, which is stored in the \memoshort.
    \item \textbf{\vlmblong (\vlmbshort)}: retrieves indices of task-relevant visual episodes from \memoshort given the current query.
    \item \textbf{\vlmclong (\vlmcshort)}: performs task classification (generate, edit, understand, or pure chat), orchestrates \vlmashort and \vlmbshort accordingly, and produces the final system output.
\end{itemize}

\noindent\textbf{Orchestration loop.} Here, an \emph{orchestration loop} denotes a single interaction iteration with the language model (i.e., one dialogue turn with ChatGPT). Within each loop, \vlmcshort (1)~classifies the incoming request to determine its task type (\emph{generation/editing} versus \emph{understanding/chat}); (2)~for tasks requiring visual context, queries \vlmbshort for relevant memory entries and fuses them with the dialogue history and current instruction into a unified context; and (3)~dispatches accordingly---rewriting the context into a refined, model-ready prompt for the image-generation/execution model (e.g., Qwen-Image-Edit~\cite{wu2025qwenimagetechnicalreport}) for generation/editing, or directly generating a text response from the unified context for understanding/chat.

To limit computational cost, only the retrieved subset of historical images is forwarded to the executor in each orchestration loop. This bounds the number of visual tokens processed per loop and nearly halves per-turn inference time (a $1.8\times$ speedup) compared to all-context baselines (see Sec.~\ref{subsec:ablation}).

\subsection{Optimizing memory construction and retrieval}
\label{subsec:learning}

\begin{figure*}[t]
  \centering
  \IfFileExists{image/data_pipeline_new.png}{\includegraphics[width=0.98\textwidth]{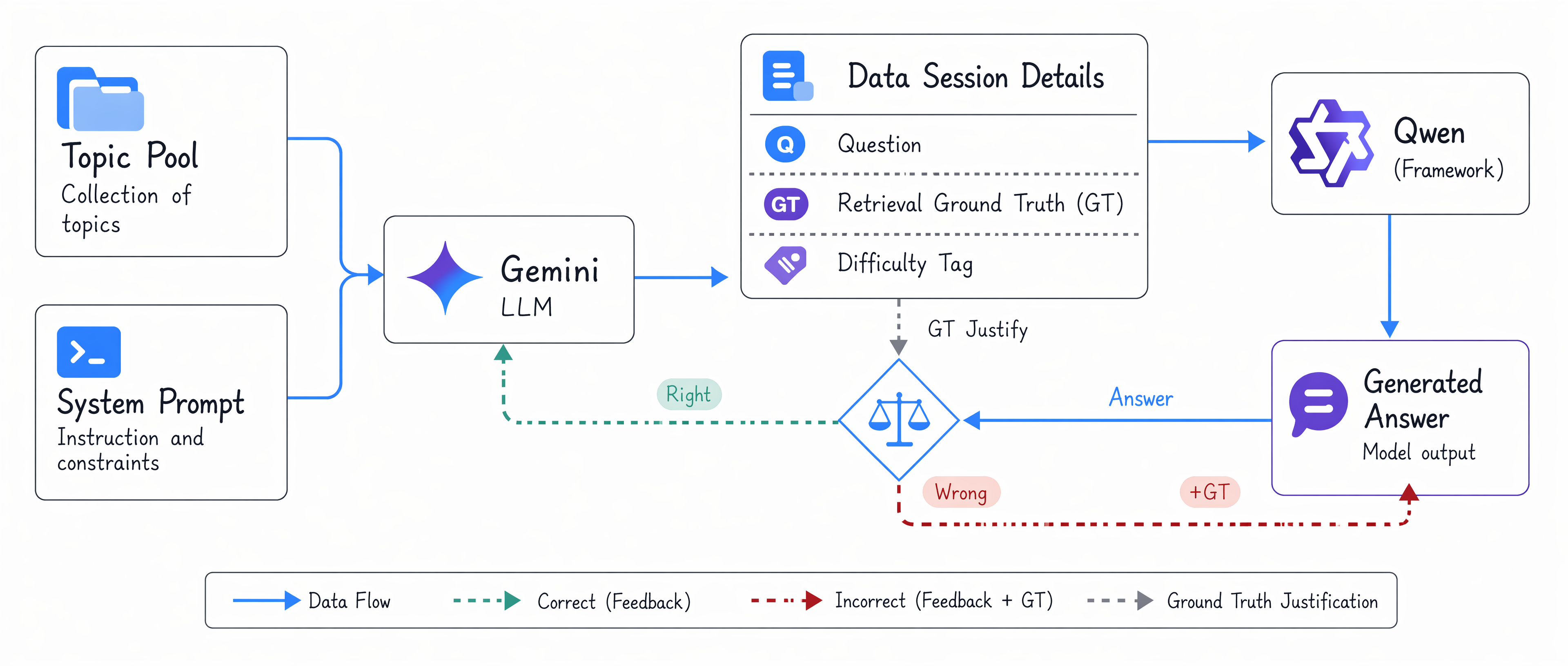}}{\fbox{\parbox{0.92\textwidth}{Data-pipeline placeholder: image/data\_pipeline\_new.png is not included in the source tree.}}}
  \caption{Closed-loop \dataengine for structured multi-turn scenario construction. A Gemini-based user simulator samples a topic and emits the next query together with its ground-truth retrieval set and difficulty tag; a zero-shot multi-agent system (Qwen-VL + Qwen-Image-Edit) produces the candidate answer; a \emph{GT-justify} verification step checks retrieval correctness and corrects mismatches before the verified answer is fed back to the simulator for the next turn. The resulting sessions are split into disjoint training and held-out (\benchshort) partitions by random seed and topic.}
  \label{fig:data_pipeline}
\end{figure*}

Accurate retrieval of relevant historical images is critical for downstream performance. We jointly optimize \vlmashort (memory construction) and \vlmbshort (memory retrieval) for retrieval success, rather than standalone captioning quality. Training data is exclusively drawn from the \dataengine pool (Sec.~\ref{subsec:data_construction}), while the evaluation benchmark \benchshort is strictly held out.

\noindent\textbf{Supervised fine-tuning of the retriever.}
Each SFT sample pairs memory entries (tags, descriptions, thumbnails) with a user query; the target is the ground-truth retrieval set. We use 300 dialogue sessions (20 turns each), split 90:10 for training and validation.

\noindent\textbf{Reinforcement learning for the retriever (DAPO).}
We apply DAPO~\cite{Yu2025DAPO} to further optimize retrieval policies. The reward is the difficulty-weighted Jaccard similarity between predicted and ground-truth retrieval sets:
\[
r = \frac{|\hat{R} \cap R^*|}{|\hat{R} \cup R^*|} \cdot w_d,
\]
where $\hat{R}$ is the prediction, $R^*$ the ground truth, and $w_d = 1.2$ for \texttt{hard}/\texttt{very\_hard} turns ($1.0$ otherwise). By convention we set $r=1$ when $R^*=\hat{R}=\varnothing$ (correctly abstaining from retrieval on a negative sample) and $r=0$ when exactly one of $R^*$, $\hat{R}$ is empty. Soft length penalties discourage malformed outputs. RL samples are filtered to emphasize (i) incorrect baseline predictions, (ii) \texttt{hard}/\texttt{very\_hard} turns, and (iii) negative samples requiring no retrieval, mitigating shortcut learning.

\noindent\textbf{Reinforcement learning for the memory writer (DAPO).}
Rather than optimizing traditional captioning metrics (CIDEr, BERTScore), we directly reward memory entries based on retrieval effectiveness. Specifically, \vlmbshort is frozen, and retrieval accuracy on memory entries produced by \vlmashort serves as the reward:
\[
r = \text{Jaccard}(\hat{R}, R^*) -  \mathcal{L}_{\text{format}}.
\]
$\mathcal{L}_{\text{format}}$ represents the format punishment. This closes the loop, ensuring memory representations are optimized for end-task utility.

\begin{table}[t]
  \centering
  \caption{Retrieval difficulty taxonomy.}
  \label{tab:difficulty}
  \small
  \renewcommand{\arraystretch}{1.15}
  \begin{tabularx}{\linewidth}{@{}l >{\raggedright\arraybackslash}X >{\raggedright\arraybackslash}p{2.8cm}@{}}
    \toprule
    \textbf{Difficulty} & \textbf{Characteristic Patterns} & \textbf{Typical Tags} \\
    \midrule
    \texttt{easy} &
    Same-topic sequential generation or editing. &
    --- \\
    \texttt{medium} &
    High-similarity variants; simple topic switch. &
    \texttt{topic\_switch} \\
    \texttt{hard} &
    Cross-topic retrieval; long-range callback ($\geq 8$ turns). &
    \shortstack[l]{\texttt{cross\_topic}\\\texttt{long\_term\_mem}} \\
    \texttt{very\_hard} &
    Multi-image comparison; fusion edits; ambiguous references. &
    \shortstack[l]{\texttt{multi\_image}\\\texttt{ambiguous\_ref}} \\
    \bottomrule
  \end{tabularx}
\end{table}

\subsection{Training Data \& Benchmark Construction}
\label{subsec:data_construction}

Existing long-horizon multimodal dialogue datasets lack turn-level retrieval supervision. We therefore build a \dataengine that programmatically generates structured 20-turn sessions with dense retrieval annotations (Fig.~\ref{fig:data_pipeline}).

The pipeline operates as a closed loop: (1)~A human-simulation module (Gemini) samples a topic, generates a query, its ground-truth retrieval set, and a difficulty tag; (2)~A zero-shot multi-agent system (Qwen-VL + Qwen-Image-Edit) produces a candidate answer; (3)~A \emph{GT-justify} mechanism verifies retrieval correctness; mismatches are corrected; (4)~The verified answer is fed back to Gemini to generate the next turn.

Table~\ref{tab:data_summary} shows the data splits of the training data and \benchshort. Training and evaluation sessions are produced with disjoint random seeds and non-overlapping topics; we denote the held-out evaluation set as \textbf{\benchlong (\benchshort)}.

\noindent\textbf{Scenario representation and topic ontology.}
Each session is structured as:
\[
S = \{(u_i, \tau_i, R_i^*, d_i, \mathbf{f}_i)\}_{i=1}^T,
\]
where $u_i$ is the user input, $\tau_i$ the task type, $R_i^*$ the ground-truth retrieval set, $d_i$ the difficulty, and $\mathbf{f}_i$ encodes retrieval challenge tags.

We curate 55 topics across eight domains (commercial, industrial, educational, public service, hospitality, natural landscape, scientific, and space); each topic defines \texttt{generate}, \texttt{edit}, \texttt{cross-reference-edit}, and \texttt{understand} prompts to ensure compositionality.
Gemini~\cite{team2023gemini} synthesizes dialogue flows with structural inductive biases (\eg, ``at least 3 long-range callbacks with span $\geq 8$ turns''), ensuring compositional and controllable conversations.

\begin{table}[t]
  \centering
  \caption{Data splits derived from the \dataengine.}
  \label{tab:data_summary}
  \small
  \resizebox{\linewidth}{!}{%
  \begin{tabular}{@{}llrl@{}}
    \toprule
    \textbf{Split} & \textbf{Scale} & \textbf{Turns} & \textbf{Usage} \\
    \midrule
    SFT train & 270 sessions $\times$ 20 turns & $\sim$5{,}400 & Retrieval SFT \\
    SFT val & 30 sessions $\times$ 20 turns & $\sim$600 & Validation \\
    RL & Filtered hard subset & varies & DAPO training \\
    \benchshort & 100 sessions $\times$ 20 turns & 2{,}000 & Evaluation \\
    \bottomrule
  \end{tabular}
  }
\end{table}

\noindent\textbf{Retrieval difficulty and hard-negative design.}
Turns are stratified into four difficulty levels based on topic shift, temporal span, multi-image interaction, and ambiguity (Table~\ref{tab:difficulty}). To prevent shortcut learning, we include (i) high-similarity confounders (near-duplicate images differing subtly) and (ii) negative retrieval samples (semantic negatives requiring no retrieval, and structural negatives prompting new generation).

\noindent\textbf{Held-out benchmark.}
The held-out benchmark consists of 100 sessions $\times$ 20 turns, generated with disjoint seeds and topics. Annotation fidelity is ensured via LLM validation and manual review. Evaluation metrics are detailed in Sec.~\ref{subsec:eval_protocol}.

\subsection{Implementation Details}
\label{subsec:implementation_details}

\vlmashort writes each image as a structured JSON memory entry (thumbnail + tags + description), while \vlmbshort operates over aligned thumbnails and memory entries and returns an image-index list. \vlmcshort classifies each turn into one of five modes: generation, editing, composition, understanding, or pure-chat, and dispatches accordingly. Text-to-image generation uses Qwen-Image, and image editing and multi-image composition use Qwen-Image-Edit; inference runs 50 denoising steps with a classifier-free guidance scale of 4.0. Full module interfaces, prompt formats, and data-construction configurations are provided in Appendix~\ref{sec:appendix_impl_details} and Appendix~\ref{sec:appendix_prompt_templates}.

\section{\cmaharness: A Tool-Augmented Engineering Deployment Framework}
\label{sec:deployed_agent}

We introduce \cmaharness, a tool-augmented deployment of the same cognitive-structured architecture in Fig.~\ref{fig:pipeline}. Rather than replacing the \agentlong framework, it extends the response-and-tool-usage branch while preserving the division of labor among \vlmashort, \memoshort, \vlmbshort, and \vlmcshort. The benchmark decomposition already works well when the user's intent is grounded in dialogue-local visual history; practical workflows, however, often require information beyond the current session---real product images, logos, names, news events, screenshots, or web pages---and benefit from remembering user preferences and project context across conversations. \cmaharness therefore keeps the memory-centric structure, assigns tool selection and invocation to \vlmcshort as the executive controller, expands the response module with external and deterministic tools, and strengthens the memory layer for persistent multi-session interaction. We defer full deployment details---including the 17-tool MEC action space, the multi-scope persistent memory schema, and the interactive tool-augmented execution loop---to Appendix~\ref{sec:appendix_cmaharness}.

\section{Experiments}
\label{sec:experiments}

Our goal is to evaluate whether the proposed \agentlong can effectively utilize long-horizon visual context while maintaining strong generation capability. To this end, we first describe the model setup (Sec.~\ref{subsec:setup}) and evaluation criteria (Sec.~\ref{subsec:eval_protocol}). We then compare our agent with unified and modular baselines on \benchshort to measure context usage accuracy in multi-turn multimodal dialogues (Sec.~\ref{subsec:Accuracy_for_context_usage}) and analyze how improved retrieval translates into downstream generation quality (Sec.~\ref{subsec:gemini_quality}). Finally, we examine the contribution of SFT/RL training (Sec.~\ref{subsec:ablation}) and ablate memory representations (Sec.~\ref{subsec:memory_representation_ablation}); a comparison against a concurrent agent framework is provided in Appendix~\ref{subsec:concurrent_agents}.

\begin{table*}[t]
  \centering
  \scriptsize
  \caption{Pipeline retrieval results on \benchshort.}
  \label{tab:pipeline_results}
  \begin{tabular}{lcccccc}
    \toprule
    \multirow{2}{*}{\textbf{Method}}
    & \multicolumn{3}{c}{\textbf{English}}
    & \multicolumn{3}{c}{\textbf{Chinese}} \\
    \cmidrule(lr){2-4} \cmidrule(lr){5-7}
    & \textbf{Full} & \textbf{Medium} & \textbf{Hard}
    & \textbf{Full} & \textbf{Medium} & \textbf{Hard} \\
    \midrule

    Unified Model (BAGEL)
    & 23.3\% & 5.6\% & 1.9\%
    & 17.8\% & 3.2\% & 2.3\% \\

    Agent Baseline-8B (All-Context)
    & 78.9\% & 67.0\% & 59.5\%
    & 76.7\% & 66.0\% & 60.8\% \\

    Agent Baseline-32B (All-Context)
    & 81.9\% & 72.1\% & 62.0\%
    & 80.3\% & 72.5\% & 71.4\% \\

    Multi-Agent Baseline-8B
    & 81.8\% & 75.2\% & 63.6\%
    & 82.5\% & 77.6\% & 77.8\% \\

    Multi-Agent Baseline-32B
    & 83.2\% & 79.4\% & 72.1\%
    & 84.5\% & 80.0\% & \textbf{79.0}\% \\

    Ours (8B)
    & \textbf{91.4}\% & \textbf{89.4}\% & \textbf{82.0}\%
    & \textbf{89.6}\% & \textbf{85.6}\% & \textbf{79.0}\% \\

    \bottomrule
  \end{tabular}
\end{table*}

\begin{table*}[t]
  \centering
  \scriptsize
  \caption{\textbf{Gemini generation quality on \benchshort}. Image Generation and Understanding/Chat are abbreviated as Gen. and U./C., respectively. Following the five task modes defined in Sec.~\ref{subsec:implementation_details}, \texttt{composition} is folded into Edit and \texttt{pure\_chat} is folded into U./C.}
  \label{tab:gemini_main}
  \begin{tabular}{lcccccccc}
    \toprule
    \multirow{2}{*}{\textbf{Method}}
    & \multicolumn{4}{c}{\textbf{English}}
    & \multicolumn{4}{c}{\textbf{Chinese}} \\
    \cmidrule(lr){2-5} \cmidrule(lr){6-9}
    & \textbf{Overall} & \textbf{Gen.} & \textbf{Edit} & \textbf{U./C.}
    & \textbf{Overall} & \textbf{Gen.} & \textbf{Edit} & \textbf{U./C.} \\
    \midrule

    Unified Model (BAGEL)
    & 4.97 & 5.11 & 4.09 & 5.77
    & 4.77 & 5.68 & 3.39 & 4.97 \\

    Agent Baseline-8B (All-Context)
    & 7.89 & 8.85 & 6.71 & 8.05
    & 7.10 & 7.86 & 6.13 & 7.15 \\

    Agent Baseline-32B (All-Context)
    & 7.93 & 9.14 & 6.38 & 8.19
    & 7.45 & 8.31 & 6.25 & 7.53 \\

    Multi-Agent Baseline-8B
    & 8.07 & 9.04 & 7.33 & 7.75
    & 7.63 & 8.35 & 6.19 & 8.03 \\

    Multi-Agent Baseline-32B
    & 8.24 & 8.97 & \textbf{7.49} & 8.22
    & 7.91 & 9.24 & 7.03 & 7.34 \\

    Ours (8B)
    & \textbf{8.49} & \textbf{9.26} & 7.31 & \textbf{8.87}
    & \textbf{8.53} & \textbf{9.38} & \textbf{7.44} & \textbf{8.56} \\

    \bottomrule
  \end{tabular}
\end{table*}

\subsection{Experimental Setup}
\label{subsec:setup}

Our system consists of three specialized vision-language modules (\vlmashort, \vlmbshort, and \vlmcshort) together with \memoshort. \vlmashort performs visual abstraction by summarizing images into tags, description and thumbnails stored in the memory bank. 
\vlmbshort conducts reasoning over the dialogue context and retrieves the relevant images from the memory bank. Finally, \vlmcshort integrates the retrieved visual context with the current dialogue state to perform task planning and generation.

Unless otherwise specified, all three modules are instantiated using Qwen3-VL-8B~\cite{bai2025qwen3vl} as the backbone model. Image generation and editing are performed using Qwen-Image-Edit~\cite{wu2025qwenimagetechnicalreport}.

We compare our approach against three representative baselines, listed below alongside our full model:

\begin{itemize}
  \item \textbf{Unified Model.} A unified multimodal model (BAGEL~\cite{Deng2025BAGEL}) that handles the entire multi-turn dialogue within a single model.
  \item \textbf{Agent Baseline-8B/32B (All-Context).} A direct-prompting baseline that concatenates the full history (all prior images and texts) into Qwen3-VL-8B/32B-Instruct for comprehension, then feeds its output as the conditional prompt---together with all historical images---to Qwen-Image-Edit for generation. As this paradigm has no explicit retrieval mechanism, we prompt Qwen3-VL to emit target image indices only during retrieval evaluation, omitting this step in generation experiments.
  \item \textbf{Multi-Agent Baseline.} A modular architecture where \vlmashort, \vlmbshort, and \vlmcshort are instantiated with Qwen3-VL-8B/32B-Instruct but trained without our long-horizon multimodal dialogue optimization.
  \item \textbf{Ours.} Our full agent trained on \dataengine-generated long-horizon dialogues, optimized with SFT and RL to improve visual abstraction and retrieval accuracy.
\end{itemize}

\subsection{Evaluation Protocol and Metrics}
\label{subsec:eval_protocol}

We evaluate all methods on \benchshort, which contains structured multi-turn multimodal dialogues with explicit image-retrieval annotations, following the difficulty definition in Table~\ref{tab:difficulty}.

\noindent\textbf{Evaluation subsets.} We report three subsets of increasing difficulty: \textbf{Full} (all 20 turns), \textbf{Medium} (turns 11--20, emphasizing longer context), and \textbf{Hard} (the \texttt{very\_hard} subset within turns 11--20), which measure how performance degrades as retrieval distance grows.

\noindent\textbf{Metrics.} We report two complementary measures: \textbf{Context Usage Accuracy}, the turn-level exact-match accuracy of the retrieved image index; and \textbf{Generation Quality}, a per-turn 0--10 score from Gemini-3-Pro assessing faithfulness and visual correctness.

\noindent\textbf{Evaluation fairness.} All methods use identical scenario sessions, turn annotations, retrieval targets, and post-processing rules; only the context-utilization mechanism differs, so performance gaps reflect context modeling rather than pipeline implementation details.

\subsection{Context Usage Accuracy}
\label{subsec:Accuracy_for_context_usage}

Table~\ref{tab:pipeline_results} reports turn-level retrieval accuracy, from which several observations emerge. \textbf{Unified models struggle with long-horizon retrieval:} BAGEL, lacking an explicit memory mechanism, reaches only 23.3\% (EN) and 17.8\% (CN) on Full and collapses to below 3\% on Hard, showing that unified architectures cannot isolate relevant visual context once the history grows long. \textbf{Structured memory filtering improves context selection:} the Multi-Agent baseline consistently beats the All-Context baseline, with the 8B VLM gaining +2.9\% (EN) / +5.8\% (CN) on Full and larger margins on Medium (+8.2\%/+11.6\%), confirming that filtering irrelevant memories is essential in long conversations.

\textbf{Task-specific training beats parameter scaling:} enlarging the Multi-Agent baseline from 8B to 32B yields only modest gains (+1.4\% EN, +2.0\% CN on Full), whereas our SFT+RL-optimized 8B model surpasses the 32B baseline by +8.2\% (EN) and +5.1\% (CN). \textbf{Advantages increase with difficulty:} the gap over the 8B Multi-Agent baseline widens from +9.6\% on Full to +14.2\% on Medium and +18.4\% on Hard, and our Full-to-Hard drop is far smaller (9.4\% vs. 18.2\%), indicating stronger robustness in long-horizon visual recall.

\subsection{Generation Quality}
\label{subsec:gemini_quality}

Table~\ref{tab:gemini_main} reports generation quality scored by Gemini-3-Pro. Our method achieves the highest overall scores (8.49 EN / 8.53 CN), and the method ranking closely mirrors retrieval accuracy, indicating that correct visual-context retrieval remains the primary bottleneck in long-horizon multimodal generation. At the task level, \textbf{Generate} is least sensitive to retrieval ($R^{*}=\varnothing$; most agents score above 7.1), \textbf{Edit} is the hardest (typically 6.1--7.5, requiring both accurate source retrieval and faithful instruction adherence; ours is best at 7.44 CN and competitive at 7.31 EN), and \textbf{Understand/Chat} benefits most from retrieval (ours reaches 8.87 EN, +0.65 over Multi-Agent Baseline-32B). Comparing the Agent Baseline-8B (All-Context) with our method, a roughly 12\% retrieval gain translates into quality improvements of +0.60 (EN) and +1.43 (CN)---a strong multiplier effect that propagates through the entire generation pipeline.

\subsection{Training Stage Analysis}
\label{subsec:ablation}

\begin{table}[t]
  \centering
  \scriptsize
  \caption{Ablation on the staged training pipeline for memory construction and retrieval. Turn-level exact-match retrieval accuracy on the English \benchshort; Full / Medium / Hard follow the subset definitions in Sec.~\ref{subsec:eval_protocol}.}
  \label{tab:ablation_english}
  \begin{tabular}{lccc}
    \toprule
    \textbf{Method} & \textbf{Full} & \textbf{Medium} & \textbf{Hard} \\
    \midrule
    Multi-Agent Baseline-8B        & 81.8\%  & 75.2\%  & 63.6\% \\
    + SFT (\vlmbshort)              & 86.6\%  & 83.7\%  & 70.9\% \\
    + RL (\vlmbshort)               & 90.1\%  & 88.3\%  & 77.4\% \\
    + RL (\vlmashort)               & \textbf{91.4}\%  & \textbf{89.4}\%  & \textbf{82.0}\% \\
    \bottomrule
  \end{tabular}
\end{table}

Our training pipeline consists of three stages; Table~\ref{tab:ablation_english} reports Hard-subset retrieval accuracy after each. \emph{Stage 1 (SFT on \vlmbshort)} improves Hard accuracy by 7.3\%, showing that supervised fine-tuning helps the retriever extract relevant visual memories from structured dialogue data. \emph{Stage 2 (RL on \vlmbshort)} adds a further 6.5\%, refining long-horizon memory selection beyond supervised signals. \emph{Stage 3 (RL on \vlmashort)} adds another 4.6\% by enhancing visual abstraction so that the most informative representations are stored, confirming that a carefully staged SFT+RL combination unlocks long-horizon multimodal context usage.

Our decoupled architecture is also more efficient (Table~\ref{tab:efficiency_ablation}, Appendix~\ref{subsec:concurrent_agents}): by delegating memory filtering to \vlmashort and \vlmbshort, the agent processes only retrieved memories rather than the full context, nearly halving per-turn inference time (12.7s vs.\ 23.1s) while also improving generation quality (8.77 vs.\ 7.93) over the Agent Baseline-32B.

\subsection{Ablation on Memory Representations}
\label{subsec:memory_representation_ablation}

Text-only LLM memories provide useful long-term semantic state, but they are not sufficient for fine-grained visual recall. We evaluate MemoryLLM-style and ReasoningBank-style text memories by supplying agents with textual abstractions of images and asking them to retrieve the relevant historical visual episode. Both text-only strategies degrade sharply on the Hard subset (24.4\% and 39.0\%, respectively), where queries often require cross-topic retrieval, long-range callbacks, or subtle visual discrimination. In contrast, our full \memoshort retains compressed thumbnails together with textual metadata and reaches 82.0\% (Table~\ref{tab:llm_memory}, Appendix~\ref{subsec:concurrent_agents}).

\section{Conclusion}

We presented a cognitive-structured multimodal agent that externalizes visual history into episodic memory and selectively retrieves relevant episodes for long-horizon dialogue. With the \benchshort benchmark and a staged SFT+RL pipeline, our 8B agent reaches 91.4\% English retrieval accuracy over 20-turn sessions, surpassing 32B baselines by +8.2\% while nearly halving inference time. The same decomposition instantiates \cmaharness, a deployed tool-augmented assistant with persistent memory, web access, generation/editing/composition, and OpenAI-compatible serving. Future work includes joint optimization, open-domain extension, and richer modalities.

\clearpage
\appendix
\section{Extended Implementation Details}
\label{sec:appendix_impl_details}

This appendix expands Sec.~\ref{subsec:implementation_details} with the full module interfaces and data-construction configurations omitted from the main text.

\noindent\textbf{Structured memory format.}
For every incoming or generated image, \vlmashort writes a structured memory entry containing a compact thumbnail, semantic tags, and a natural-language description. The memory-writing prompt requires valid JSON with fields such as \texttt{tags} and \texttt{description}; this format makes the stored episode both machine-readable and interpretable. In practice, tags provide high-recall lexical anchors, while the description preserves spatial, relational, and stylistic cues that are difficult to capture with short labels alone.

\noindent\textbf{Multimodal retrieval interface.}
\vlmbshort receives a list of candidate thumbnails aligned with candidate indices, textual memory entries containing tags and descriptions, and the current user query. It outputs an image-index list, with an empty list indicating that no historical image is needed. This interface is shared by SFT, RL, and evaluation, ensuring that the model is optimized under the same input-output contract used at inference time.

\noindent\textbf{Executive controller modes.}
\vlmcshort classifies each turn into generation, editing, composition, understanding, or pure-chat modes. For generation, it rewrites short user requests into detailed visual prompts. For editing and composition, it grounds the instruction in retrieved images and emits a precise editing prompt that specifies the target region or object while preserving unrelated content. For understanding and chat, it answers directly using the retrieved visual evidence, dialogue history, and memory metadata.

\noindent\textbf{Generation and editing backbones.}
Text-to-image generation is handled by Qwen-Image, while image editing and multi-image composition use Qwen-Image-Edit. Unless otherwise stated, inference uses 50 denoising steps and a classifier-free guidance scale of 4.0. During data construction, we use a faster Qwen-Image-Edit-2511-Lightning configuration with four denoising steps to reduce generation cost; final evaluation uses the standard high-quality pipeline.

\noindent\textbf{Prompt interfaces.}
The module prompts are intentionally format-constrained so that memory writing, retrieval, task classification, prompt rewriting, and visual answering expose stable input-output contracts across SFT, RL, evaluation, and deployment. We provide the full prompt templates in Appendix~\ref{sec:appendix_prompt_templates}.

\section{Prompt Templates}
\label{sec:appendix_prompt_templates}

We provide the core prompt templates used by the agent modules to clarify the module interfaces and support reproducibility. The wording is intentionally simple and format-constrained, so that memory writing, retrieval, task classification, prompt rewriting, and visual answering expose stable input-output contracts across training, evaluation, and deployment.

\begin{tcolorbox}[agentprompt, colback=gray!5, colframe=gray!60, title=\vlmashort Prompt (Image $\rightarrow$ Memory)]
\textless image\textgreater{} Please analyze this image in detail and generate complete image features + description in English. The output must be JSON containing the following fields: tags (tag list, at least 5 tags, array format), description (detailed description, at least 100 words). Please ensure the output is valid JSON.
\end{tcolorbox}

\begin{tcolorbox}[agentprompt, colback=blue!3, colframe=blue!50, title=\vlmbshort Prompt (Retrieval)]
Candidate image list (aligned with the order in `images'):\\
- rank 0: \textless image\textgreater\\
- rank 1: \textless image\textgreater\\
\ldots\\
Memory bank information (including tags + description):\\
\{[0] | tags: sunset, sports car, highway | description: A sleek black \ldots\}\\
\{[1] | tags: beach, ocean, sunset | description: A panoramic view \ldots\}\\
\ldots\\
User query: \{q\}\\
Please select the most relevant image ID list based on the user query and memory bank information. If no historical image is needed, output an empty list.
\end{tcolorbox}

\begin{tcolorbox}[agentprompt, colback=green!3, colframe=green!50, title=\vlmcshort Prompt (Task Classification)]
You are a multimodal task classification expert. Given the user input and current state, categorize the task into one of the following five types: generate, edit, composite, understand, pure\_chat.\\
Current state: User text = \{user\_input\}; Image status = \{image\_info\}.\\
Output requirement: Return JSON only, with no explanation: \{"task\_type": "generate" | "edit" | "composite" | "understand" | "pure\_chat"\}.
\end{tcolorbox}

\begin{tcolorbox}[agentprompt, colback=orange!3, colframe=orange!50, title=\vlmcshort Prompt (Generate Mode Rewriting)]
You are a prompt rewriting assistant for text-to-image generation models. Convert a short user instruction into a detailed visual prompt. Identify the main subject, expand visual attributes such as color, shape, size, and appearance, add reasonable scene or environment details if not specified, optionally include viewpoint or composition, and keep the user's original intent unchanged. Output only the final rewritten prompt. User instruction: \{instruction\}.
\end{tcolorbox}

\begin{tcolorbox}[agentprompt, colback=red!3, colframe=red!50, title=\vlmcshort Prompt (Edit Mode Rewriting)]
You are a multimodal prompt rewriting assistant for image editing models. You will receive an image, a user editing instruction, and optional dialogue context. Identify the object or region to modify using the image, clearly describe the requested modification, preserve the user's intent, explicitly indicate that all other parts of the image should remain unchanged, and output only one concise editing prompt. Image editing instruction: \{instruction\}.
\end{tcolorbox}

\begin{tcolorbox}[agentprompt, colback=purple!3, colframe=purple!50, title=\vlmcshort Prompt (Understanding / Chat)]
Text-only mode: Current user instruction = \{user\_instruction\}; task type = pure\_chat. Please answer directly.\\
Contextual mode: \textless image\textgreater{} \textless image\textgreater{} \ldots; dialogue history; retrieved memories with tags and descriptions; current user instruction = \{user\_instruction\}; task type = mmu. Please reason carefully based on the instruction and reference images, and provide a detailed answer.
\end{tcolorbox}

\section{\cmaharness Deployment Details}
\label{sec:appendix_cmaharness}

This appendix expands the deployment overview presented in Sec.~\ref{sec:deployed_agent}. We describe how the core \agentlong is wrapped into a service-facing harness, the MEC-driven tool registry, the flexible persistent memory system, and the interactive tool-augmented execution loop.

\subsection{From Core Agent System to Deployed Harness}
\label{subsec:cmaharness_overview}

\cmaharness follows the same information flow as \agentlong. User interactions first enter a dialogue state that contains the current instruction, uploaded images, and prior turns. New visual inputs are first persisted as memory assets and then asynchronously abstracted by a \vlmashort-like card extractor into structured tags, captions, palettes, and thumbnails. These entries are written into \memoshort and later retrieved by \vlmbshort when the user refers to previous visual context. The central difference lies in the operational scope of \vlmcshort. In the original benchmark pipeline, \vlmcshort mainly decides whether the task requires understanding, retrieval, generation, or editing, and then prepares the prompt for the downstream multimodal model. In \cmaharness, \vlmcshort becomes a tool-calling executive: it still performs task decomposition and multimodal context fusion, but it can additionally call external and deterministic tools before producing the final answer.

This design keeps the cognitive interpretation of Fig.~\ref{fig:pipeline}. \vlmashort remains responsible for visual compression, including tags, captions, and thumbnails. \memoshort remains the long-horizon visual memory substrate. \vlmbshort remains the retrieval module that selects relevant memory for the current turn. \vlmcshort remains the main reasoning module, but its action space is enlarged from a small set of generation/editing choices to a structured set of callable tools. As a result, \cmaharness is best understood as a harness around \agentlong: it provides service interfaces, tool schemas, persistent storage, and execution policies around the same cognitive modules.

\subsection{MEC-Driven Action Space Expansion}
\label{subsec:cmaharness_tools}

\cmaharness exposes a registry of 17 tools to \vlmcshort. Each tool is described by a name, natural-language capability description, JSON schema, concurrency-safety flag, progress message, and asynchronous execution function. The registry is converted into an OpenAI-style function-calling interface, so \vlmcshort can reason about the user's goal, decide whether tools are necessary, choose the appropriate tool, fill its arguments, observe the result, and continue the loop if another action is needed. The tools extend the original response-and-tool-usage block in Fig.~\ref{fig:pipeline}: generation and editing tools include image generation, image editing, multi-image composition, deterministic collage construction, background removal, watermark removal, cropping, and text overlay; understanding and memory tools include image inspection, best-image selection, and image retrieval; external-information tools include web search, batch search, web-image search, web-page fetching, and image fetching.

The key design choice is that tool access is governed by an MEC action policy rather than exposed as an undifferentiated toolbox. \vlmcshort first decides whether a tool is needed at all; many requests are better answered directly without external execution. When the requested output is a purely imagined scene, \vlmcshort can call text-to-image generation. When the output must contain a specific real-world visual entity, such as a logo, product, screenshot, artwork, or named person's face, \vlmcshort is instructed not to rely on a verbal description, because text-only generation or editing will often hallucinate the identity. Instead, it first obtains real pixels through memory retrieval or web-image fetching, and then routes the task to composition or editing with those image identifiers as visual references.

\cmaharness also separates semantic composition from deterministic layout. Multi-image diffusion composition is used when several visual references must be fused into a new scene. In contrast, poster, infographic, timeline, comparison, and many-item showcase tasks often require faithful faces, logos, captions, dates, or CJK text. For these cases, \vlmcshort can choose a PIL-based collage tool that preserves each input by pixel-level placement and renders text deterministically, optionally followed by a lightweight beautification pass. This division makes the enlarged MEC action space practical: \vlmcshort can route each request to the operation that best preserves what the user cares about, instead of forcing all visual tasks through a single generative or editing model.

Tool execution is constrained by the same executive role assigned to \vlmcshort. Consecutive read-only operations, such as web search, web fetch, image inspection, and memory retrieval, can be executed concurrently to reduce latency. Mutating or GPU-heavy operations, such as image generation, editing, composition, cropping, and text rendering, are serialized to preserve memory consistency and avoid resource contention. The loop tracks candidate images produced during a turn and can use an image-selection tool to choose the best final deliverable, keeping intermediate drafts from being over-exposed to the user. To prevent tool use from degenerating into uncontrolled loops, \cmaharness adds simple controller-level constraints. If \vlmcshort repeatedly issues the same tool call with the same arguments, the system prompts it to revise its plan instead of executing the duplicate call indefinitely. Each user turn also has a maximum number of tool rounds; when this budget is exhausted, \vlmcshort summarizes the current progress and returns a graceful partial response. These mechanisms keep tool invocation under MEC control while still allowing multi-step workflows.

\subsection{Flexible Persistent Multimodal Memory}
\label{subsec:cmaharness_memory}

\cmaharness extends \memoshort from a session-local visual memory bank into a flexible persistent memory system. The \agentlong focuses on controlled memory construction and retrieval within a fixed dialogue, whereas deployment requires memory to support multiple users, multiple sessions, reusable visual references, long transcripts, generated variants, fetched assets, and user feedback. We therefore organize memory into several scopes: session memory for the current interaction, user-level memory for durable preferences and project facts, gallery memory for reusable visual assets, and compact summaries for long-running conversations.

At the visual level, each image is stored as a three-part asset. The original image is preserved for future editing, composition, inspection, and download; a compact thumbnail is used for efficient retrieval and UI preview; and a JSON image card stores semantic tags, a short description, color palette, aspect ratio, file paths, and lifecycle metadata. The lifecycle fields are important in deployed workflows: parent-child links track edited variants, an ``is current'' flag records the active deliverable, user feedback marks preferred or rejected outputs, and selected images can be promoted into a reusable gallery. This turns \vlmashort's output from a transient caption into a structured memory object that can be updated and reused across turns.

Beyond visual assets, \cmaharness maintains user-scope textual memories. These memories are stored as typed Markdown entries, including user preferences, feedback, project facts, and external references, with an automatically refreshed memory index. At prompt construction time, the memory manager assembles session summaries, user memories, and the visual memory index into a compact context block for \vlmcshort. This allows MEC to condition tool selection and response generation not only on the current dialogue, but also on persistent user intent and project-level constraints.

Retrieval is adaptive rather than uniform. Obvious references such as ``the latest image'', ``the first image'', or explicit image indices are resolved by deterministic rules without invoking a large model. When many candidates exist, a lightweight text prefilter narrows the candidate set using tags, descriptions, palette, recency, and feedback. The final selection uses multimodal retrieval over both image-card metadata and thumbnails, allowing \vlmbshort to disambiguate references that are difficult to capture textually, such as ``the blue one'' or ``the busier layout''. Full-resolution pixels are loaded only after a target image is selected and a downstream tool requires them.

The memory layer is also maintained asynchronously so that memory quality can improve without blocking the foreground interaction. When a new image is uploaded, generated, edited, fetched, or composed, \cmaharness immediately stores the original image and thumbnail and can return the result to the user at once. A background extractor then analyzes the thumbnail and enriches the corresponding image card with semantic tags, a concise description, and color-palette information. This separates latency-sensitive interaction from slower memory annotation: the user does not wait for full visual abstraction, but subsequent turns can still benefit from richer retrieval cues.

Long conversations are handled similarly. Rather than replaying an unbounded transcript into \vlmcshort, \cmaharness compacts older turns into structured summaries once a session exceeds token or tool-call thresholds, while keeping the most recent turns verbatim. The summary preserves high-level goals, established preferences, generated image identifiers, open follow-ups, and the current active state. Thus, long-term context remains available to the controller without continuously increasing the prompt size. Together, these mechanisms make \memoshort in \cmaharness more flexible than the benchmark memory bank: it is multi-scope, multi-type, lifecycle-aware, and persistent, yet still preserves the token-efficient retrieval principle of the original \agentlong.

\subsection{Interactive Tool-Augmented Agent Execution}
\label{subsec:cmaharness_deployment}

Finally, \cmaharness turns the cognitive pipeline into an interactive execution loop. For each user turn, the system reconstructs the current memory context, rebuilds the system prompt with the latest memory blocks and tool schemas, lets \vlmcshort reason over the task, and executes any selected tools. Tool results are appended back into the transcript and then fed to \vlmcshort for the next decision, so the controller can iteratively search, retrieve, inspect, generate, edit, or compose until it has enough information to answer. This event-driven design keeps the foreground interaction responsive: intermediate reasoning, tool steps, tool results, images, and final text can be surfaced progressively while the same underlying MEC loop remains responsible for the complete decision process.

The deployed instance uses Qwen3.5-122B-A10B as the MEC planner, Qwen-Image-2512 for text-to-image generation, and Qwen-Image-Edit-2511 for editing and multi-image composition. These model choices are implementation instantiations of the modules rather than changes to the architecture. To make the agent practical for multi-step use, the planner and image backbones are kept as persistent components instead of being reloaded for every request. The generation and editing paths also include automatic prompt engineering: generation prompts are expanded toward richer visual descriptions, while editing prompts emphasize locality, object grounding, and preservation of unrelated regions.

The interaction loop also preserves the agent's stateful multimodal semantics across turns. Uploaded or produced images are stored as addressable memory assets, tool outputs are recorded as part of the dialogue state, and the current deliverable is tracked so follow-up instructions such as ``make it brighter'' or ``use the previous poster'' can be resolved without reattaching every historical image. Lightweight safeguards, including duplicate-call detection, per-turn tool-round limits, and graceful partial summaries, keep the loop bounded while still allowing long multi-step workflows. Together, these choices turn the original \agentlong design into a practical \cmaharness: the same PAE--EVM--CoRE--MEC structure, but with a broader MEC action space, stronger persistent memory, and interactive execution for open-ended multimodal workflows.

\section{Additional Experimental Analysis}
\label{subsec:concurrent_agents}

\noindent\textbf{Efficiency analysis.} Table~\ref{tab:efficiency_ablation} compares our decoupled design with the Agent Baseline-32B, which processes the entire multimodal history in a single model. By delegating memory filtering to \vlmashort and \vlmbshort, the agent retrieves only relevant visual memories instead of the full context, reducing the visual tokens fed into the generation module. As a result, per-turn inference time is nearly halved (12.7s vs.\ 23.1s) while generation quality also increases (8.77 vs.\ 7.93).

\begin{table}[t]
  \centering
  \scriptsize
  \caption{Ablation study showing the efficiency and quality gains from the proposed \vlmashort+\vlmbshort modules.}
  \label{tab:efficiency_ablation}
  \resizebox{\linewidth}{!}{%
  \begin{tabular}{lcccc}
    \toprule
    \textbf{Method} & \textbf{Full} & \textbf{Hard} & \textbf{Quality} & \textbf{Runtime (s)} \\
    \midrule
    Agent-Baseline-32B             & 81.9\% & 62.0\% & 7.93 & 23.1 \\
    \vlmashort+\vlmbshort + Qwen3VL-32B           & \textbf{92.0}\% & \textbf{83.5}\% & \textbf{8.77} & \textbf{12.7} \\
    \bottomrule
  \end{tabular}
  }
\end{table}

\noindent\textbf{Comparison with concurrent agent frameworks.} We further compare our agent with GenArtist~\cite{wang2024genartist}, a concurrent multimodal agent framework for unified image generation and editing. To expose long-horizon drift, we split English \benchshort sessions into early (turns 1--10) and late (turns 11--20) stages and separately evaluate generation and editing quality. Table~\ref{tab:agent_comparison} shows that GenArtist degrades substantially in late turns, while our agent remains stable. This result supports the central claim that tool orchestration alone is insufficient: without explicit episodic visual memory and retrieval, multi-turn generation and editing suffer from context confusion as sessions grow.

\begin{table}[t]
\centering
\scriptsize
\caption{Multi-turn generation and editing quality scores (0--10) on English \benchshort. Sessions are split into early and late stages to highlight robustness against long-horizon semantic drift.}
\label{tab:agent_comparison}
\begin{tabular}{@{}lcccc@{}}
\toprule
\multirow{2}{*}{\textbf{Method}} & \multicolumn{2}{c}{\textbf{Generation Quality}} & \multicolumn{2}{c}{\textbf{Editing Quality}} \\
\cmidrule(lr){2-3} \cmidrule(l){4-5}
 & \textbf{Early} & \textbf{Late} & \textbf{Early} & \textbf{Late} \\
\midrule
GenArtist & 6.77 & 4.80 & 5.29 & 4.03 \\
\textbf{Ours (8B Agent)} & \textbf{9.51} & \textbf{9.06} & \textbf{7.74} & \textbf{7.13} \\
\bottomrule
\end{tabular}
\end{table}

\noindent\textbf{Memory-representation ablation.} Table~\ref{tab:llm_memory} details the memory-representation ablation summarized in Sec.~\ref{subsec:memory_representation_ablation}, comparing text-only LLM memory systems against our full \memoshort across difficulty subsets.

\begin{table}[t]
\centering
\scriptsize
\caption{Ablation on \benchshort comparing text-only LLM memory systems against our full \memoshort. Retrieval accuracy (\%) is reported across difficulty subsets.}
\label{tab:llm_memory}
\begin{tabular}{@{}lccc@{}}
\toprule
\textbf{Memory Strategy} & \textbf{Full} & \textbf{Medium} & \textbf{Hard} \\
\midrule
MemoryLLM-style memory & 51.7\% & 47.2\% & 24.4\% \\
ReasoningBank-style memory & 68.2\% & 51.8\% & 39.0\% \\
\midrule
\textbf{Ours (Full \memoshort)} & \textbf{91.4\%} & \textbf{89.4\%} & \textbf{82.0\%} \\
\bottomrule
\end{tabular}
\end{table}

\section{Additional Qualitative Examples}
\label{sec:appendix_qualitative_examples}

We include additional long-horizon dialogue examples to illustrate how episodic visual memory supports mixed generation, editing, composition, and visual question answering. One example begins with a Mars habitat, later edits it with a dust storm, and then correctly answers questions about earlier visual details such as solar panels. A second example performs environmental transformations such as converting a sunny street scene into a rainy night version while preserving spatial layout, then recalls distant details from a separate savanna scene. A third example spans indoor scenes, logistics environments, wildlife landscapes, and Mars habitats, requiring object insertion, cross-scene composition, and detailed visual comparison. Across these examples, the agent demonstrates long-range visual recall, topic switching without memory loss, multi-image composition, and grounded multimodal reasoning.

\begin{figure*}[t]
  \centering
  \IfFileExists{image/case_1410.pdf}{\includegraphics[width=\textwidth]{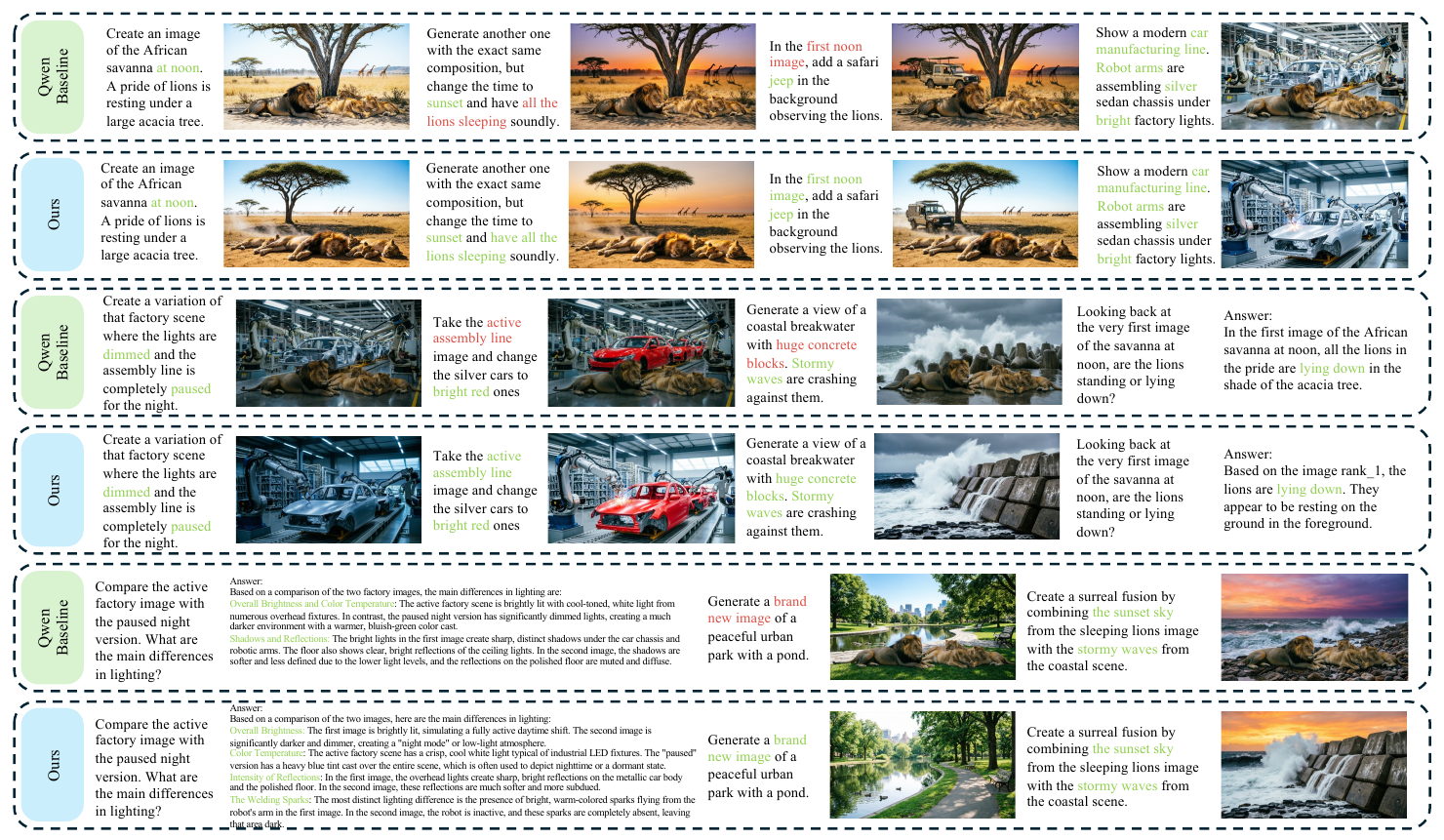}}{\fbox{\parbox{0.92\textwidth}{Qualitative comparison placeholder: image/case\_1410.pdf is not included in the source tree.}}}
  \caption{Qualitative comparison between the baseline and our method on multi-turn visual understanding, generation, and editing.}
  \label{fig:case}
\end{figure*}

We first contrast our method with the all-context baseline (Fig.~\ref{fig:case}). In the all-context baseline, lack of filtering causes hallucination: the agent over-relies on the first image and incorrectly propagates its content into subsequent generations. In contrast, our staged training ensures accurate retrieval and, coupled with \vlmcshort's prompt rewriting, produces outputs that adhere closely to user instructions across multi-turn interactions.

\begin{figure*}[t]
\centering
\IfFileExists{image/case1.pdf}{\includegraphics[width=\linewidth]{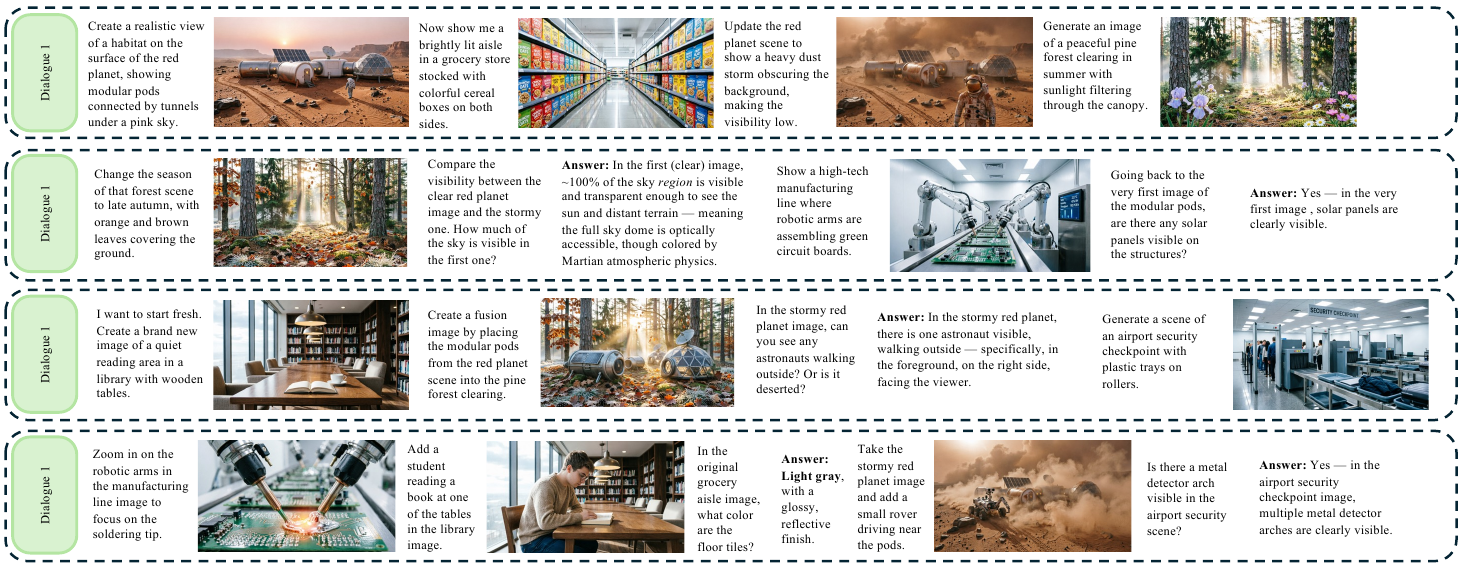}}{\fbox{\parbox{0.92\linewidth}{Additional qualitative example 1 placeholder: case1.pdf is not included in the source tree.}}}
\caption{\textbf{Example Dialogue 1.} A long-horizon multimodal interaction containing image generation, editing, topic switching, and visual question answering.}
\label{fig:case1}
\end{figure*}

\begin{figure*}[t]
\centering
\IfFileExists{image/case2.pdf}{\includegraphics[width=\linewidth]{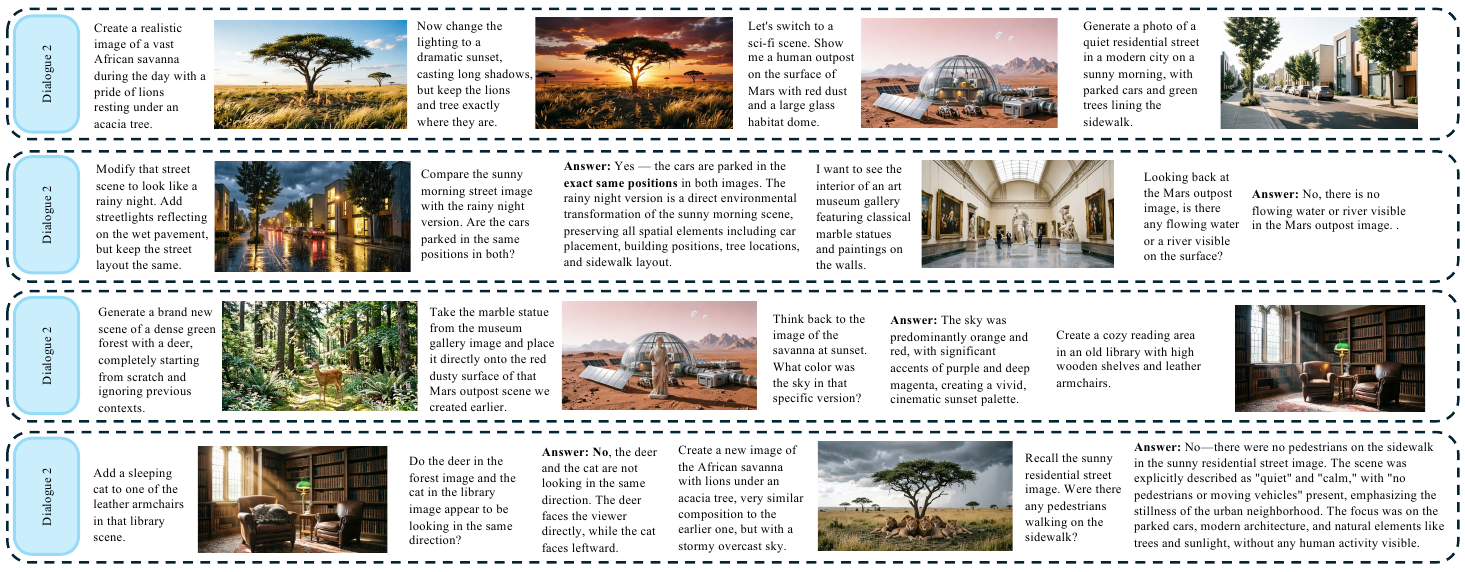}}{\fbox{\parbox{0.92\linewidth}{Additional qualitative example 2 placeholder: case2.pdf is not included in the source tree.}}}
\caption{\textbf{Example Dialogue 2.} A dialogue emphasizing environmental transformations, long-range recall, and stable visual consistency.}
\label{fig:case2}
\end{figure*}

\begin{figure*}[t]
\centering
\IfFileExists{image/case3.pdf}{\includegraphics[width=\linewidth]{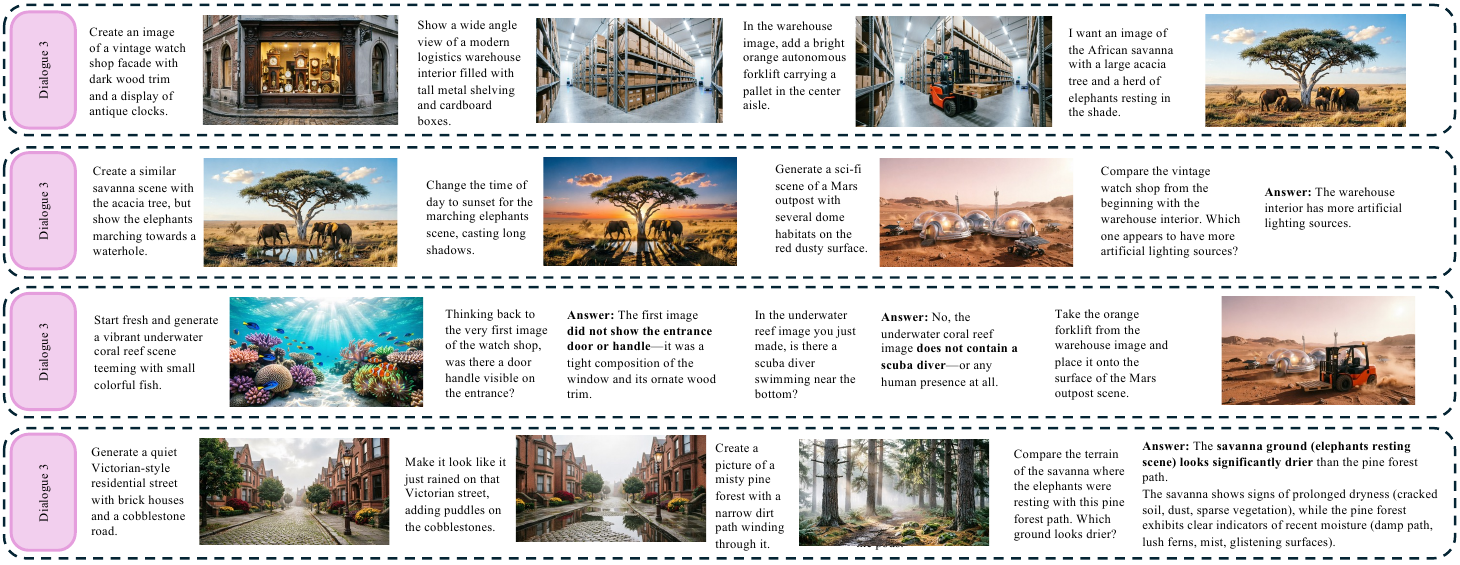}}{\fbox{\parbox{0.92\linewidth}{Additional qualitative example 3 placeholder: case3.pdf is not included in the source tree.}}}
\caption{\textbf{Example Dialogue 3.} A complex interaction involving object insertion, cross-scene composition, and detailed visual comparison across distant turns.}
\label{fig:case3}
\end{figure*}

{\small
\bibliographystyle{ieeenat_fullname}
\bibliography{main}

@article{Yu2025DAPO,
  title={DAPO: An Open-Source LLM Reinforcement Learning System},
  author={Yu, Qiying and others},
  journal={arXiv preprint arXiv:2503.14476},
  year={2025}
}

@article{schick2024toolformer,
  title={Toolformer: Language models can teach themselves to use tools},
  author={Schick, Timo and Dwivedi-Yu, Jane and Dess{\`\i}, Roberto and Raileanu, Roberta and Lomeli, Maria and Zettlemoyer, Luke and Cancedda, Nicola and Scialom, Thomas},
  journal={Advances in Neural Information Processing Systems},
  volume={36},
  year={2024}
}

@inproceedings{Li2025SPORT,
  title={Iterative Trajectory Exploration for Multimodal Agents},
  author={Li, Pengxiang and others},
  booktitle={Advances in Neural Information Processing Systems},
  year={2025}
}

@article{Jiang2026SYNAPSE,
  title={SYNAPSE: Empowering LLM Agents with Episodic-Semantic Memory via Spreading Activation},
  author={Jiang, Hanqi and others},
  journal={arXiv preprint arXiv:2601.02744},
  year={2026}
}

@article{Yeo2025WorldMM,
  title={WorldMM: Dynamic Multimodal Memory Agent for Long Video Reasoning},
  author={Yeo, Woongyeong and others},
  journal={arXiv preprint arXiv:2512.02425},
  year={2025}
}

@inproceedings{fan2024videoagent,
  title={Videoagent: A memory-augmented multimodal agent for video understanding},
  author={Fan, Yue and Ma, Xiaojian and Wu, Rujie and Du, Yuntao and Li, Jiaqi and Gao, Zhi and Li, Qing},
  booktitle={European Conference on Computer Vision},
  pages={75--92},
  year={2024},
  organization={Springer}
}

@article{wang2024genartist,
  title={Genartist: Multimodal llm as an agent for unified image generation and editing},
  author={Wang, Zhenyu and Li, Aoxue and Li, Zhenguo and Liu, Xihui},
  journal={Advances in Neural Information Processing Systems},
  volume={37},
  pages={128374--128395},
  year={2024}
}

@inproceedings{li2023camel,
  title={Camel: Communicative agents for" mind" exploration of large language model society},
  author={Li, Guohao and Hammoud, Hasan Abed Al Abbas and Itani, Hani and Khizbullin, Dmitrii and Ghanem, Bernard},
  booktitle={Advances in Neural Information Processing Systems},
  volume={36},
  year={2023}
}

@inproceedings{Venkatesh2025CREA,
  title={CREA: A Collaborative Multi-Agent Framework for Creative Content Generation with Diffusion Models},
  author={Venkatesh, Kavana and others},
  booktitle={Advances in Neural Information Processing Systems},
  year={2025}
}

@inproceedings{li2023blip,
  title={Blip-2: Bootstrapping language-image pre-training with frozen image encoders and large language models},
  author={Li, Junnan and Li, Dongxu and Savarese, Silvio and Hoi, Steven},
  booktitle={International conference on machine learning},
  pages={19730--19742},
  year={2023},
  organization={PMLR}
}

@article{ChameleonTeam2024,
  title={Chameleon: Mixed-Modal Early-Fusion Foundation Models},
  author={{Chameleon Team}},
  journal={arXiv preprint arXiv:2405.09818},
  year={2024}
}

@article{wang2024emu3,
  title={Emu3: Next-Token Prediction is All You Need},
  author={Wang, Xinlong and Zhang, Xiaosong and Luo, Zhengxiong and Sun, Quan and Cui, Yufeng and Wang, Jinsheng and Zhang, Fan and Wang, Yueze and Li, Zhen and Yu, Qiying and others},
  journal={arXiv preprint arXiv:2409.18869},
  year={2024}
}

@article{Deng2025BAGEL,
  title={Emerging Properties in Unified Multimodal Pretraining},
  author={Deng, Chaorui and others},
  journal={arXiv preprint arXiv:2505.14683},
  year={2025}
}

@article{Xie2025Showo2,
  title={Show-o2: Improved Native Unified Multimodal Models},
  author={Xie, Jinheng and others},
  journal={arXiv preprint arXiv:2506.15564},
  year={2025}
}

@inproceedings{Ma2025JanusFlow,
  title={JanusFlow: Harmonizing Autoregression and Rectified Flow for Unified Multimodal Understanding and Generation},
  author={Ma, Yiyang and others},
  booktitle={Proceedings of the IEEE/CVF Conference on Computer Vision and Pattern Recognition},
  year={2025}
}

@inproceedings{brooks2023instructpix2pix,
  title={Instructpix2pix: Learning to follow image editing instructions},
  author={Brooks, Tim and Holynski, Aleksander and Efros, Alexei A},
  booktitle={Proceedings of the IEEE/CVF Conference on Computer Vision and Pattern Recognition},
  pages={18392--18402},
  year={2023}
}

@article{Hu2026TalkPhoto,
  title={TalkPhoto: A Versatile Training-Free Conversational Assistant for Intelligent Image Editing},
  author={Hu, Yujie and others},
  journal={arXiv preprint arXiv:2601.01915},
  year={2026}
}

@inproceedings{Zhou2025MultiTurn,
  title={Multi-turn Consistent Image Editing},
  author={Zhou, Zijun and others},
  booktitle={Proceedings of the IEEE/CVF International Conference on Computer Vision},
  year={2025}
}

@inproceedings{Wang2024MMNeedle,
  title={Multimodal Needle in a Haystack},
  author={Wang, Hengyi and others},
  booktitle={Proceedings of the 2025 Conference of the North American Chapter of the Association for Computational Linguistics: Human Language Technologies},
  year={2025}
}

@article{Wu2024VisualHaystacks,
  title={Visual Haystacks: A Vision-Centric Needle-In-A-Haystack Benchmark},
  author={Wu, Tsung-Han and others},
  journal={arXiv preprint arXiv:2407.13766},
  year={2024}
}

@misc{wu2025qwenimagetechnicalreport,
      title={Qwen-Image Technical Report}, 
      author={Chenfei Wu and Jiahao Li and Jingren Zhou and Junyang Lin and Kaiyuan Gao and Kun Yan and Sheng-ming Yin and Shuai Bai and Xiao Xu and Yilei Chen and Yuxiang Chen and Zecheng Tang and Zekai Zhang and Zhengyi Wang and An Yang and Bowen Yu and Chen Cheng and Dayiheng Liu and Deqing Li and Hang Zhang and Hao Meng and Hu Wei and Jingyuan Ni and Kai Chen and Kuan Cao and Liang Peng and Lin Qu and Minggang Wu and Peng Wang and Shuting Yu and Tingkun Wen and Wensen Feng and Xiaoxiao Xu and Yi Wang and Yichang Zhang and Yongqiang Zhu and Yujia Wu and Yuxuan Cai and Zenan Liu},
      year={2025},
      eprint={2508.02324},
      archivePrefix={arXiv},
      primaryClass={cs.CV},
      url={https://arxiv.org/abs/2508.02324}, 
}

@inproceedings{huang2025wegen,
  title={Wegen: A unified model for interactive multimodal generation as we chat},
  author={Huang, Zhipeng and Zhuang, Shaobin and Fu, Canmiao and Yang, Binxin and Zhang, Ying and Sun, Chong and Zhang, Zhizheng and Wang, Yali and Li, Chen and Zha, Zheng-Jun},
  booktitle={Proceedings of the Computer Vision and Pattern Recognition Conference},
  pages={23679--23689},
  year={2025}
}

@article{achiam2023gpt,
  title={Gpt-4 technical report},
  author={Achiam, Josh and Adler, Steven and Agarwal, Sandhini and Ahmad, Lama and Akkaya, Ilge and Aleman, Florencia Leoni and Almeida, Diogo and Altenschmidt, Janko and Altman, Sam and Anadkat, Shyamal and others},
  journal={arXiv preprint arXiv:2303.08774},
  year={2023}
}

@article{touvron2023llama,
  title={Llama: Open and efficient foundation language models},
  author={Touvron, Hugo and Lavril, Thibaut and Izacard, Gautier and Martinet, Xavier and Lachaux, Marie-Anne and Lacroix, Timoth{\'e}e and Rozi{\`e}re, Baptiste and Goyal, Naman and Hambro, Eric and Azhar, Faisal and others},
  journal={arXiv preprint arXiv:2302.13971},
  year={2023}
}

@article{bai2023qwen,
  title={Qwen technical report},
  author={Bai, Jinze and Bai, Shuai and Chu, Yunfei and Cui, Zeyu and Dang, Kai and Deng, Xiaodong and Fan, Yang and Ge, Wenbin and Han, Yu and Huang, Fei and others},
  journal={arXiv preprint arXiv:2309.16609},
  year={2023}
}

@article{qwen3,
    title={Qwen3 Technical Report}, 
    author={An Yang and Anfeng Li and Baosong Yang and Beichen Zhang and Binyuan Hui and Bo Zheng and Bowen Yu and Chang Gao and Chengen Huang and Chenxu Lv and Chujie Zheng and Dayiheng Liu and Fan Zhou and Fei Huang and Feng Hu and Hao Ge and Haoran Wei and Huan Lin and Jialong Tang and Jian Yang and Jianhong Tu and Jianwei Zhang and Jianxin Yang and Jiaxi Yang and Jing Zhou and Jingren Zhou and Junyang Lin and Kai Dang and Keqin Bao and Kexin Yang and Le Yu and Lianghao Deng and Mei Li and Mingfeng Xue and Mingze Li and Pei Zhang and Peng Wang and Qin Zhu and Rui Men and Ruize Gao and Shixuan Liu and Shuang Luo and Tianhao Li and Tianyi Tang and Wenbiao Yin and Xingzhang Ren and Xinyu Wang and Xinyu Zhang and Xuancheng Ren and Yang Fan and Yang Su and Yichang Zhang and Yinger Zhang and Yu Wan and Yuqiong Liu and Zekun Wang and Zeyu Cui and Zhenru Zhang and Zhipeng Zhou and Zihan Qiu},
    journal = {arXiv preprint arXiv:2505.09388},
    year={2025}
}

@misc{liu2024llavanext,
    title={LLaVA-NeXT: Improved reasoning, OCR, and world knowledge},
    url={https://llava-vl.github.io/blog/2024-01-30-llava-next/},
    author={Liu, Haotian and Li, Chunyuan and Li, Yuheng and Li, Bo and Zhang, Yuanhan and Shen, Sheng and Lee, Yong Jae},
    month={January},
    year={2024}
}

@article{bai2025qwen3vl,
  title={Qwen3-vl technical report},
  author={Bai, Shuai and Cai, Yuxuan and Chen, Ruizhe and Chen, Keqin and Chen, Xionghui and Cheng, Zesen and Deng, Lianghao and Ding, Wei and Gao, Chang and Ge, Chunjiang and others},
  journal={arXiv preprint arXiv:2511.21631},
  year={2025}
}

@article{wang2025internvl3,
  title={Internvl3. 5: Advancing open-source multimodal models in versatility, reasoning, and efficiency},
  author={Wang, Weiyun and Gao, Zhangwei and Gu, Lixin and Pu, Hengjun and Cui, Long and Wei, Xingguang and Liu, Zhaoyang and Jing, Linglin and Ye, Shenglong and Shao, Jie and others},
  journal={arXiv preprint arXiv:2508.18265},
  year={2025}
}

@article{peng2023kosmos,
  title={Kosmos-2: Grounding multimodal large language models to the world},
  author={Peng, Zhiliang and Wang, Wenhui and Dong, Li and Hao, Yaru and Huang, Shaohan and Ma, Shuming and Wei, Furu},
  journal={arXiv preprint arXiv:2306.14824},
  year={2023}
}

@article{zhu2023minigpt,
  title={Minigpt-4: Enhancing vision-language understanding with advanced large language models},
  author={Zhu, Deyao and Chen, Jun and Shen, Xiaoqian and Li, Xiang and Elhoseiny, Mohamed},
  journal={arXiv preprint arXiv:2304.10592},
  year={2023}
}

@article{jin2025videomem,
  title={VideoMem: Enhancing Ultra-Long Video Understanding via Adaptive Memory Management},
  author={Jin, Hongbo and Wang, Qingyuan and Zhang, Wenhao and Liu, Yang and Cheng, Sijie},
  journal={arXiv preprint arXiv:2512.04540},
  year={2025}
}

@article{ouyang2025reasoningbank,
  title={Reasoningbank: Scaling agent self-evolving with reasoning memory},
  author={Ouyang, Siru and Yan, Jun and Hsu, I and Chen, Yanfei and Jiang, Ke and Wang, Zifeng and Han, Rujun and Le, Long T and Daruki, Samira and Tang, Xiangru and others},
  journal={arXiv preprint arXiv:2509.25140},
  year={2025}
}

@article{ye2025agentfold,
  title={AgentFold: Long-Horizon Web Agents with Proactive Context Management},
  author={Ye, Rui and Zhang, Zhongwang and Li, Kuan and Yin, Huifeng and Tao, Zhengwei and Zhao, Yida and Su, Liangcai and Zhang, Liwen and Qiao, Zile and Wang, Xinyu and others},
  journal={arXiv preprint arXiv:2510.24699},
  year={2025}
}

@article{long2025seeing,
  title={Seeing, listening, remembering, and reasoning: A multimodal agent with long-term memory},
  author={Long, Lin and He, Yichen and Ye, Wentao and Pan, Yiyuan and Lin, Yuan and Li, Hang and Zhao, Junbo and Li, Wei},
  journal={arXiv preprint arXiv:2508.09736},
  year={2025}
}

@inproceedings{wang2024videoagent,
  title={Videoagent: Long-form video understanding with large language model as agent},
  author={Wang, Xiaohan and Zhang, Yuhui and Zohar, Orr and Yeung-Levy, Serena},
  booktitle={European Conference on Computer Vision},
  pages={58--76},
  year={2024},
  organization={Springer}
}

@article{kumar2024mmctagent,
  title={Mmctagent: Multi-modal critical thinking agent framework for complex visual reasoning},
  author={Kumar, Somnath and Gadhia, Yash and Ganu, Tanuja and Nambi, Akshay},
  journal={arXiv preprint arXiv:2405.18358},
  year={2024}
}

@article{wang2025yanyun,
  title={Yanyun-3: Enabling Cross-Platform Strategy Game Operation with Vision-Language Models},
  author={Wang, Guoyan and Huang, Yanyan and Chen, Chunlin and Wang, Lifeng and Sun, Yuxiang},
  journal={arXiv preprint arXiv:2511.12937},
  year={2025}
}

@article{liu2025agent0,
  title={Agent0-vl: Exploring self-evolving agent for tool-integrated vision-language reasoning},
  author={Liu, Jiaqi and Xiong, Kaiwen and Xia, Peng and Zhou, Yiyang and Ji, Haonian and Feng, Lu and Han, Siwei and Ding, Mingyu and Yao, Huaxiu},
  journal={arXiv preprint arXiv:2511.19900},
  year={2025}
}

@article{gao2024multi,
  title={Multi-modal agent tuning: Building a vlm-driven agent for efficient tool usage},
  author={Gao, Zhi and Zhang, Bofei and Li, Pengxiang and Ma, Xiaojian and Yuan, Tao and Fan, Yue and Wu, Yuwei and Jia, Yunde and Zhu, Song-Chun and Li, Qing},
  journal={arXiv preprint arXiv:2412.15606},
  year={2024}
}

@inproceedings{hong2023metagpt,
  title={MetaGPT: Meta programming for a multi-agent collaborative framework},
  author={Hong, Sirui and Zhuge, Mingchen and Chen, Jonathan and Zheng, Xiawu and Cheng, Yuheng and Wang, Jinlin and Zhang, Ceyao and Wang, Zili and Yau, Steven Ka Shing and Lin, Zijuan and others},
  booktitle={The twelfth international conference on learning representations},
  year={2023}
}

@inproceedings{qian2024chatdev,
  title={Chatdev: Communicative agents for software development},
  author={Qian, Chen and Liu, Wei and Liu, Hongzhang and Chen, Nuo and Dang, Yufan and Li, Jiahao and Yang, Cheng and Chen, Weize and Su, Yusheng and Cong, Xin and others},
  booktitle={Proceedings of the 62nd annual meeting of the association for computational linguistics (volume 1: Long papers)},
  pages={15174--15186},
  year={2024}
}

@inproceedings{Bai2023QwenVLAV,
  title={Qwen-VL: A Versatile Vision-Language Model for Understanding, Localization, Text Reading, and Beyond},
  author={Jinze Bai and Shuai Bai and Shusheng Yang and Shijie Wang and Sinan Tan and Peng Wang and Junyang Lin and Chang Zhou and Jingren Zhou},
  year={2023},
  url={https://api.semanticscholar.org/CorpusID:261101015}
}

@article{tang2025unilip,
  title={Unilip: Adapting clip for unified multimodal understanding, generation and editing},
  author={Tang, Hao and Xie, Chenwei and Bao, Xiaoyi and Weng, Tingyu and Li, Pandeng and Zheng, Yun and Wang, Liwei},
  journal={arXiv preprint arXiv:2507.23278},
  year={2025}
}

@article{team2023gemini,
  title={Gemini: a family of highly capable multimodal models},
  author={Team, Gemini and Anil, Rohan and Borgeaud, Sebastian and Alayrac, Jean-Baptiste and Yu, Jiahui and Soricut, Radu and Schalkwyk, Johan and Dai, Andrew M and Hauth, Anja and Millican, Katie and others},
  journal={arXiv preprint arXiv:2312.11805},
  year={2023}
}
}
\end{document}